\documentclass[conference]{IEEEtran}
\IEEEoverridecommandlockouts

\usepackage{cite}
\usepackage{amsmath,amssymb,amsfonts}
\usepackage{algorithmic}
\usepackage{graphicx}
\usepackage{textcomp}
\usepackage{xcolor}
\usepackage{multirow}
\usepackage{hyperref}
\usepackage[top=0.75in, bottom=1in, left=0.625in, right=0.625in]{geometry}

\def\BibTeX{{\rm B\kern-.05em{\sc i\kern-.025em b}\kern-.08em
    T\kern-.1667em\lower.7ex\hbox{E}\kern-.125emX}}
\begin{document}

\title{A Fusion-Guided Inception Network for Hyperspectral Image Super-Resolution\\
}

\author{
    \IEEEauthorblockN{
        Usman Muhammad\textsuperscript{1} and Jorma Laaksonen\textsuperscript{1}
    }
    \IEEEauthorblockA{\textsuperscript{1} Department of Computer Science, Aalto University, Finland}
}


\maketitle

\begin{abstract}
The fusion of low-spatial-resolution hyperspectral images (HSIs) with high-spatial-resolution conventional images (e.g., panchromatic or RGB) has played a significant role in recent advancements in HSI super-resolution. However, this fusion process relies on the availability of precise alignment between image pairs, which is often challenging in real-world scenarios. To mitigate this limitation, we propose a single-image super-resolution model called the Fusion-Guided Inception Network (FGIN). Specifically, we first employ a spectral–spatial fusion module to effectively integrate spectral and spatial information at an early stage. Next, an Inception-like hierarchical feature extraction strategy is used to capture multiscale spatial dependencies, followed by a dedicated multi-scale fusion block. To further enhance reconstruction quality, we incorporate an optimized upsampling module that combines bilinear interpolation with depthwise separable convolutions. Experimental evaluations on two publicly available hyperspectral datasets demonstrate the competitive performance of our method. The source codes are publicly available at: \href{https://github.com/Usman1021/fusion}{https://github.com/Usman1021/fusion}.
\end{abstract}


\begin{figure*}[htbp]
    \centering
    \includegraphics[width=0.95\textwidth]{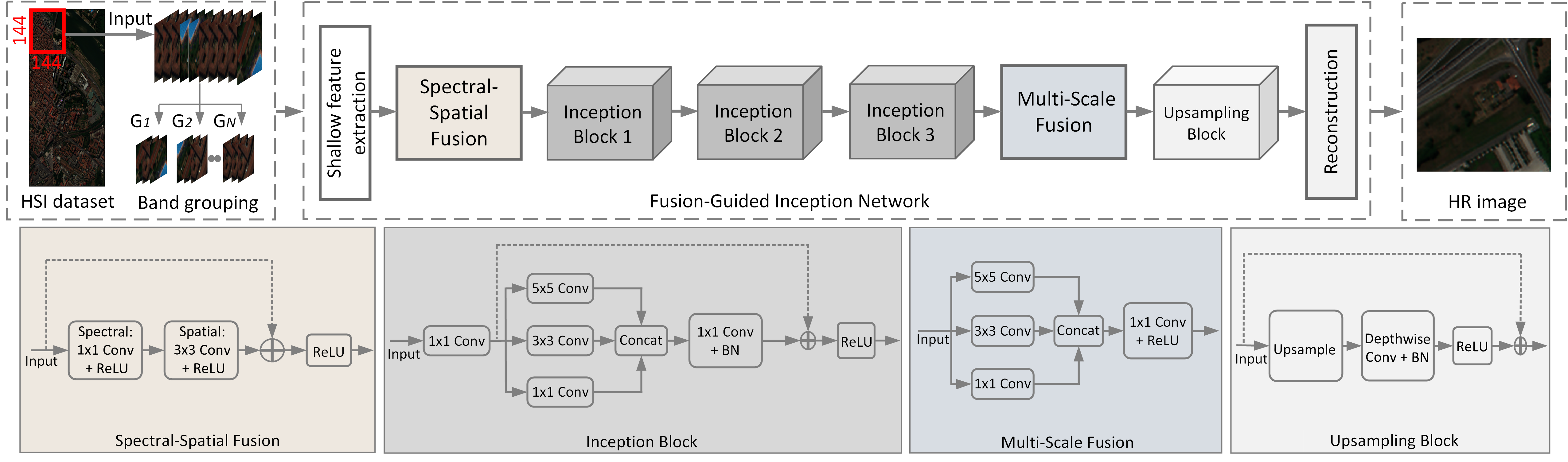} 
\caption{Overview of the proposed FGIN model. The architecture consists of: (1) a Spectral-Spatial Fusion module (left); (2) Inception-like multiscale feature extractors (center) for hierarchical representation learning; (3) a Multi-Scale Fusion block for feature refinement; and (4) an Upsampling block (right) that reconstructs high-resolution (HR) hyperspectral images.}
    \label{fig}
\end{figure*}

\section{Introduction}
Hyperspectral images (HSIs) consist of hundreds of distinct, discrete spectral bands spanning wavelengths from the visible to the infrared spectrum (400–2500 nm). This broad range of spectral information makes HSIs highly valuable for a wide range of applications, including surveillance, medical image processing, and satellite scene classification \cite{peng2022lcrca, muhammad2018pre, muhammad2018feature, muhammad2019bag, muhammad2022patch}. However, environmental factors (e.g., clouds, haze), the density of the sensor array, and limitations on sensor size make it difficult to directly acquire images with both high spatial and high spectral resolution \cite{wang2024single}. As a result, hyperspectral image super-resolution (SR) has become a critical research focus, aiming to enhance image quality and extract finer details from captured data.

The main objective of single-image super-resolution is to enhance a degraded low-resolution (LR) image by reconstructing its corresponding high-resolution (HR) version \cite{li2019hyperspectral}. Since panchromatic images offer higher spatial resolution and contain richer details, many hyperspectral SR techniques leverage panchromatic, RGB, or multispectral images to assist the reconstruction process, thereby enhancing the details of low-resolution hyperspectral images \cite{wang2024single}. Currently, hyperspectral image super-resolution (HSI SR) methods can be broadly categorized into two approaches based on the use of auxiliary images: fusion-based hyperspectral image SR and single-image super-resolution (SISR) \cite{yang2025deep}.

Fusion-based hyperspectral image super-resolution enhances spatial detail by integrating a hyperspectral image with high-resolution images of the same scene. Techniques such as Bayesian inference \cite{wei2015bayesian}, matrix decomposition \cite{yokoya2011coupled}, and sparse representation \cite{akhtar2014sparse} have demonstrated strong performance in this area. However, these methods typically assume precise alignment between the low-resolution hyperspectral image and the high-resolution auxiliary image, which can be difficult to achieve in real-world scenarios \cite{zhou2019integrated}.
In contrast, single-image hyperspectral super-resolution methods aim to reconstruct high-resolution hyperspectral images directly from low-resolution inputs without relying on auxiliary data \cite{muhammad2025towards}. For example, Yang et al. \cite{yang2023hierarchical} proposed a dual-pyramid model that progressively estimates high-resolution images by hierarchically fusing spatial and spectral information through a pre-trained Laplacian pyramid network. Similarly, Zhang et al. \cite{zhang2024hyperspectral} introduced a feature diversity loss function to improve single-image super-resolution performance.

In addition, conventional super-resolution methods, such as interpolation-based, regularization-based, and sparse representation approaches, have been widely utilized in earlier studies\cite{xu2024hyperspectral}. In particular, interpolation-based methods often degrade performance because of their inability to recover high-frequency details, while regularization-based approaches rely on handcrafted priors and are computationally intensive. Similarly, sparse representation methods suffer from high complexity and may introduce artifacts due to patch-wise processing \cite{xu2024hyperspectral}. All of these conventional methods struggle to generalize across diverse image content. Moreover, due to the lack of auxiliary information, single-image hyperspectral super-resolution relies on extracting high-frequency details within a limited spatial dimension, often leading to a decrease in reconstruction performance compared to fusion-based methods \cite{wu2023combining}. Furthermore, downsampling of the image causes the loss of high-frequency components such as textures, sharp edges, and fine structures, making single-image super-resolution an ill-posed problem.

To mitigate the aforementioned challenges, we propose a single-image super-resolution model that enhances image reconstruction through spectral–spatial fusion, multi-scale feature extraction, and optimized upsampling. Specifically, we begin with a shallow feature extractor (i.e., a 3×3 convolutional layer), followed by spectral–spatial fusion \cite{zhong2017spectral} and Inception-like blocks \cite{szegedy2015going} to capture diverse contextual features. Motivated by this, we incorporate a multi-scale fusion block to further refine the integration by emphasizing both fine-grained spectral characteristics and local spatial coherence. Finally, we apply an optimized upsampling strategy that leverages bilinear interpolation, followed by depthwise separable convolutions \cite{howard2017mobilenets} and residual skip connections \cite{he2016deep}, to preserve fine details.

In summary, our key contributions are threefold:
\begin{enumerate}
\item We propose a novel single-image super-resolution model called the Fusion-Guided Inception Network (FGIN), which leverages spectral–spatial fusion and integrates Inception-like hierarchical feature extraction to enhance spatial detail while maintaining spectral consistency in hyperspectral images.
\item We design an optimized upsampling block that combines bilinear interpolation, depthwise convolutions, and residual skip connections to effectively preserve both spatial detail and spectral integrity.
\item We conduct experiments under 2$\times$, 4$\times$, and 8$\times$ downsampling scenarios, demonstrating highly competitive performance on two publicly available hyperspectral datasets.
\end{enumerate}

\section{Methodology}
Fig. 1 presents an overview of our proposed Fusion-Guided Inception Network (FGIN), which consists of four key components: (1) spectral–spatial fusion, (2) Inception-like blocks, (3) multi-scale fusion, and (4) an upsampling module. We begin with a description of the band grouping process, followed by a detailed analysis of each component in the subsequent subsections.

\subsection{Band Grouping}

Hyperspectral images typically contain hundreds of spectral bands, making it computationally intensive to process all bands simultaneously. Moreover, the high spectral correlation between bands can introduce redundancy. To address these issues, we adopt a band grouping strategy \cite{wang2024enhancing}, which partitions adjacent bands into overlapping groups for efficient integration with our model. Specifically, the spectral bands are divided into overlapping subgroups by specifying a fixed group size and an overlap parameter, ensuring that consecutive groups share common bands to preserve spectral continuity.

\subsection{Spectral-Spatial Fusion}

We begin with shallow feature extraction using a \(3 \times 3\) convolution with 32 filters and ReLU activation, followed by the spectral-spatial fusion block. Unlike parallel-branch designs, this fusion module operates sequentially. First, a \(1 \times 1\) pointwise convolution is applied to extract spectral (channel-wise) correlations:

\begin{equation}
H_{\text{s}} = \sigma(W_{1 \times 1} * X + b_s)
\end{equation}
where \(X \in \mathbb{R}^{H \times W \times C}\) represents the input feature map, \(W_{1 \times 1}\) denotes the spectral convolution weights, \(b_s\) is the corresponding bias, and \(\sigma(\cdot)\) is the ReLU activation function. Next, the intermediate spectral features \(H_{\text{s}}\) are passed through a \(3 \times 3\) convolution to capture local spatial structure:

\begin{equation}
H_{\text{sp}} = \sigma(W_{3 \times 3} * H_{\text{s}} + b_p)
\end{equation}
where \(W_{3 \times 3}\) and \(b_p\) represent the spatial convolution weights and bias, respectively. To integrate these features with the original input, a residual connection is applied:

\begin{equation}
Y = \sigma(H_{\text{sp}} + X)
\end{equation}
where \(Y\) is the residual-enhanced output feature map that combines spectral features extracted by a \(1 \times 1\) convolution (\(H_{\text{s}}\)), spatial features from a \(3 \times 3\) convolution applied to \(H_{\text{s}}\) (\(H_{\text{sp}}\)), and input features added via residual connection (\(X\)).

\begin{table}
\caption{Quantitative results and model complexity}
\label{tab:fgin_model_comparison}
\centering
\resizebox{0.9\columnwidth}{!}{
\begin{tabular}{c|c|c}
\hline
\multicolumn{3}{c}{Ablation study on PaviaU (4×)} \\
\hline
Model Variant & MPSNR$\uparrow$ & SAM$\downarrow$ \\
\hline
FGIN w/o band grouping          & 29.33  & 5.970 \\
\hline
FGIN w/s 16                     & 29.38  & 3.331 \\
\hline
FGIN  w/s 32                  & 30.33  & 4.819 \\
\hline
FGIN w/s 48                     & 30.18  & 5.965 \\
\hline
FGIN w/o spectral-fusion    & 30.49  & 4.864 \\
\hline
FGIN + bilinear upsampling    & 30.44 & 4.961 \\
\hline
\multicolumn{3}{c}{} \\ 
\hline
\multicolumn{3}{c}{Model Complexity} \\
\hline
Model & Scale & Parameters \\
\hline
ERCSR \cite{li2021exploring}         & 4 & 1.59M \\
\hline
MCNet \cite{li2020mixed}             & 4 & 2.17M \\
\hline
PDENet \cite{hou2022deep}            & 4 & 2.30M \\
\hline
CSSFENet \cite{zhang2024hyperspectral} & 4 & 1.61M \\
\hline
FGIN (Ours)                           & 4 & \textbf{1.07M} \\
\hline
\end{tabular}
}
\end{table}

\subsection{Inception-like Blocks}

To capture diverse spatial features at multiple receptive fields, we empirically include three Inception-like blocks inspired by the Inception architecture \cite{szegedy2015going}, combined with residual learning \cite{he2016deep}. Let $X \in \mathbb{R}^{H \times W \times C}$ denote the input feature map, where $H$, $W$, and $C$ represent the height, width, and number of channels, respectively. Each branch computes its feature map as:

\begin{equation}
\begin{aligned}
F_{i} &= \sigma(\text{BN}(W_{i} * X + b_{i})), \\
i &\in \{\text{1$\times$1},\ \text{1$\times$1$\rightarrow$3$\times$3},\ \text{1$\times$1$\rightarrow$5$\times$5},\ \text{1$\times$1}\}
\end{aligned}
\end{equation}
where $W_{i}$ and $b_{i}$ are the convolutional weights and biases for the $i$-th branch, $\sigma(\cdot)$ denotes the ReLU activation function, and BN represents batch normalization.

The outputs of the four branches are concatenated along the channel dimension and passed through a $1 \times 1$ convolutional layer to project the merged features back to the original dimensionality. A residual connection is then added by summing the projected features with the input $X$, followed by a ReLU activation:

\begin{equation}
F_{\text{out}} = \sigma(\text{BN}(W_{f} * [F_1, F_2, F_3, F_4]) + X)
\end{equation}
where $W_{f}$ is the $1 \times 1$ convolution used to fuse the concatenated output, and $[F_1, F_2, F_3, F_4]$ denotes channel-wise concatenation. This structure allows the model to effectively learn multi-scale spatial features while preserving training stability and improving gradient flow through residual learning.

\subsection{Multi-Scale Fusion Block}
Due to the presence of both fine and coarse structures across spectral bands, we employ a multi-scale fusion block to extract and integrate spatial features at multiple receptive fields. Given an input feature map \(x\), the block applies three parallel convolutional operations with kernel sizes of \(1{\times}1\), \(3{\times}3\), and \(5{\times}5\) to capture local, medium, and global spatial information, respectively: $b_1 = \sigma(\text{Conv}_{1{\times}1}(x))$, $b_2 = \sigma(\text{Conv}_{3{\times}3}(x))$, $b_3 = \sigma(\text{Conv}_{5{\times}5}(x))$. These outputs are concatenated along the channel dimension to form a multi-scale representation:
\begin{equation}
M_{\text{c}} = [b_1, b_2, b_3]
\end{equation}
To fuse the aggregated features and project them back to the original channel dimension, a final \(1{\times}1\) convolution is applied:
\begin{equation}
L_{\text{d}} = \sigma(\text{Conv}_{1{\times}1}(M_{\text{c}}))
\end{equation}
where \(L_{\text{d}}\) represents the final output feature map.

\begin{table*}
\centering
\caption{Evaluation on datasets (PaviaC, PaviaU) in different scaling setups. The comparison results are reported from \cite{zhang2024hyperspectral}.}
\resizebox{0.79\textwidth}{!}{%
\renewcommand{\arraystretch}{0.85} 
\small 
\begin{tabular}{|c|c|ccc|ccc|}
\hline
\multirow{2}{*}{\textbf{Scale Factor}} & \multirow{2}{*}{\textbf{Model}} & \multicolumn{3}{c|}{\textbf{PaviaC}} & \multicolumn{3}{c|}{\textbf{PaviaU}} \\ 
\cline{3-8}
                      &                   & \textbf{MPSNR$\uparrow$} & \textbf{MSSIM$\uparrow$} & \textbf{SAM$\downarrow$} & \textbf{MPSNR$\uparrow$} & \textbf{MSSIM$\uparrow$} & \textbf{SAM$\downarrow$} \\ \hline
\multirow{8}{*}{\centering $\boldsymbol{2\times}$} 
    & VDSR   \cite{kim2016accurate}     & 34.87 & 0.9501 & 3.689  & 34.03  & 0.9524 & 3.258 \\ 
    & EDSR   \cite{lim2017enhanced}    & 34.58 & 0.9452 & 3.898  & 33.98  & 0.9511 & 3.334 \\ 
    & MCNet   \cite{li2020mixed}    & 34.62 & 0.9455 & 3.865  & 33.74  & 0.9502 & 3.359 \\ 
    & MSDformer    \cite{chen2023msdformer}    & 35.02 & 0.9493 & 3.691  & 34.15  & 0.9553 & 3.211 \\ 
    & MSFMNet  \cite{zhang2021multi}           & 35.20 & 0.9506 & 3.656  & 34.98  & 0.9582 & 3.160 \\ 
    & AS3 ITransUNet    \cite{xu20233}   & 35.22 & 0.9511 & 3.612  & 35.16  & 0.9591 & 3.149\\ 
    & PDENet    \cite{hou2022deep}      & 35.24 & 0.9519 & \underline{3.595}  & 35.27  & \underline{0.9594} & \underline{3.142} \\ 
    & CSSFENet  \cite{zhang2024hyperspectral} & \underline{35.52} & \underline{0.9544} & \textbf{3.542}  & \underline{35.92}  & \textbf{0.9625}  & \textbf{3.038} \\ 
    & FGIN (Ours)    & \textbf{36.57} &  \textbf{0.9570} & 3.737 & \textbf{35.95} &  0.9439 & 3.768 \\ \hline
\multirow{8}{*}{\centering $\boldsymbol{4\times}$} 
    & VDSR \cite{kim2016accurate}      & 28.31     &0.7707  &6.514  & 29.90   &0.7753  &4.997 \\ 
    & EDSR  \cite{lim2017enhanced}    & 28.59  &0.7782  &6.573  & 29.89   & 0.7791 &5.074 \\ 
    & MCNet   \cite{li2020mixed}   & 28.75   &0.7826  &6.385  & 29.99  &0.7835  &4.917 \\ 
    & MSDformer   \cite{chen2023msdformer}      & 28.81  &0.7833  &5.897  & 30.09  & 0.7905  &4.885 \\ 
    & MSFMNet  \cite{zhang2021multi}        &   28.87  & 0.7863  &6.300  & 30.28  &0.7948 &4.861 \\ 
    & AS3 ITransUNet  \cite{xu20233}  &  28.87  &0.7893  &5.972  & 30.28  & 0.7940  &4.859\\ 
    & PDENet  \cite{hou2022deep}       & 28.95  &0.7900  &5.876  & 30.29  & 0.7944 & 4.853 \\ 
    & CSSFENet   \cite{zhang2024hyperspectral}   & \underline{29.05} & \underline{0.7961} & \underline{5.816}  & \textbf{30.68} & \textbf{0.8107} & \underline{4.839} \\ 
    & FGIN  (Ours)       & \textbf{29.58} & \textbf{0.8140} & \textbf{4.875} & \underline{30.33} & \underline{0.7979} & \textbf{4.819} \\ \hline
\multirow{8}{*}{\centering $\boldsymbol{8\times}$} 
    & VDSR  \cite{kim2016accurate}     & 24.80   &0.4944  &7.588  & 27.02  &0.5962 &7.133 \\ 
    & EDSR   \cite{lim2017enhanced}   & 25.06   &0.5282  &7.507 & 27.46  &0.6302 &6.678 \\ 
    & MCNet  \cite{li2020mixed}    & 25.09   &0.5391&7.429  & 27.48   &0.6254   &6.683 \\ 
    & MSDformer  \cite{chen2023msdformer}       & 25.21   &0.5462   &7.427  & 27.32  &0.6341 &6.668 \\ 
    & MSFMNet  \cite{zhang2021multi}        &   25.25   &0.5464  &7.449  & 27.58  &0.6356 &6.615 \\ 
    & AS3 ITransUNet  \cite{xu20233}  &  25.25    &0.5435&7.417  & 27.68  &0.6413 &6.574\\ 
    & PDENet   \cite{hou2022deep}      & 25.28   &0.5436  &7.402  & 27.73  & \underline{0.6457} & 6.531 \\ 
    & CSSFENet   \cite{zhang2024hyperspectral}   & \underline{25.35} & \underline{0.5493} & \underline{7.306}   & \underline{27.82} & \textbf{0.6569} & \underline{6.505} \\ 
    & FGIN (Ours)        & \textbf{25.75} & \textbf{0.5815} & \textbf{6.419} & \textbf{27.92} & 0.6242 & \textbf{6.233} \\ \hline
\end{tabular}%
}
\label{tab:results_paviac_paviau_centered_algorithm_scalefactor}
\end{table*}

\subsection{Upsampling Block}

To enhance spatial resolution, we adopt an optimized upsampling strategy that combines bilinear interpolation with depthwise separable convolutions~\cite{howard2017mobilenets}, followed by a residual connection~\cite{he2016deep}. Let \( X \in \mathbb{R}^{H \times W \times C} \) denote the input low-resolution feature map. The upsampling process begins with bilinear interpolation, expressed as:
\begin{equation}
\hat{X} = \mathcal{U}(X, s)
\end{equation}
where \( \mathcal{U} \) denotes bilinear interpolation with scale factor \( s \), and \( \hat{X} \) is the upsampled feature map. Next, depthwise convolution is applied to refine the spatial structure:
\begin{equation}
D = \sigma(\text{BN}(W_d * \hat{X} + b_d))
\end{equation}
where \( W_d \) and \( b_d \) represent the depthwise convolution kernel and its corresponding bias, \( \text{BN}(\cdot) \) denotes batch normalization, and \( \sigma(\cdot) \) is the ReLU activation function. To retain low-frequency contextual information, a residual shortcut path is constructed by directly upsampling the original input:
\begin{equation}
X_{\text{res}} = \mathcal{U}(X, s)
\end{equation}

The final output is obtained by element-wise summation of the main and residual paths:
\begin{equation}
X_{\text{out}} = D + X_{\text{res}}
\end{equation}
where \( X_{\text{out}} \) is the output feature map that integrates both learned spatial details and preserved contextual information.

\section{Experimental setup}

\subsection{Implementation}
Two publicly available hyperspectral datasets, PaviaC and PaviaU, containing 102 and 103 spectral bands respectively, are used in this study. Following the experimental setup described in \cite{zhang2024hyperspectral}, we adopt a patch-based training and testing strategy with a patch size of $144 \times 144$ pixels. Specifically, test patches are extracted from the bottom center of the PaviaC image and the top left corner of the PaviaU image, while the remaining image regions are used for training \cite{zhang2024hyperspectral}.

To generate low-resolution inputs, the extracted patches are downsampled using area-based interpolation with scale factors of 2×, 4×, and 8×. The model is trained using the Adam optimizer with a batch size of 4. An early stopping criterion is applied to prevent overfitting and to eliminate the need for a fixed number of training epochs. For band grouping, both datasets are divided into spectral groups of size 32, with one-fourth overlap between adjacent groups to ensure spectral continuity and consistency. To quantitatively evaluate reconstruction quality, we adopt three widely used metrics: mean peak signal-to-noise ratio (MPSNR), mean structural similarity index (MSSIM), and spectral angle mapper (SAM) \cite{chudasama2024comparison}.

\subsection{Ablation Study}
Table I presents the ablation results, highlighting the contributions of band grouping and the impact of model complexity in the FGIN model. It can be seen that removing the band grouping strategy results in a noticeable performance drop (MPSNR: 29.33, SAM: 5.970), confirming its importance for effective spectral reconstruction. Experiments with varying band group sizes reveal that a group size of 16 achieves the best SAM value (3.33), indicating better spectral fidelity, while the configuration with group size 32 yields the highest MPSNR (30.33), balancing both spatial and spectral reconstruction quality. Interestingly, when the spectral-spatial fusion block is removed, MPSNR slightly increases to 30.49, but SAM also increases to 4.864, suggesting that this component plays a crucial role in preserving spectral consistency. 

Additionally, replacing the optimized upsampling block with a bilinear only (i.e., without depthwise convolution) achieves a comparable MPSNR (30.44) but results in a degraded SAM (4.961), again indicating a loss in spectral accuracy. In general, these findings validate the effectiveness of the design choices of FGIN to maintain a favorable balance between performance and model efficiency.

\subsection{Comparison with State-of-the-Art Methods}
As presented in Table I, we compare the parameter count of our proposed FGIN model with several state-of-the-art hyperspectral super-resolution methods, including ERCSR (1.59M) \cite{li2021exploring}, MCNet (2.17M) \cite{li2020mixed}, PDENet (2.30M) \cite{hou2022deep}, and CSSFENet (1.61M) \cite{zhang2024hyperspectral}. In contrast, FGIN comprises only 1.07 million parameters, significantly reducing model size while maintaining competitive reconstruction performance. 

Table II presents a detailed quantitative comparison under multiple scaling factors (2×, 4×, and 8×) on both the PaviaC and PaviaU datasets. The comparison includes various state-of-the-art methods such as VDSR \cite{kim2016accurate}, EDSR \cite{lim2017enhanced}, MCNet \cite{li2020mixed}, MSDformer \cite{chen2023msdformer}, MSFMNet \cite{zhang2021multi}, AS3 ITransUNet \cite{xu20233}, PDENet \cite{hou2022deep}, and CSSFENet \cite{zhang2024hyperspectral}. Despite its lightweight nature, our model ranks first in the PaviaC dataset and achieves competitive performance on the PaviaU dataset in terms of PSNR and SAM. This highlights the efficiency of FGIN in balancing model complexity and accuracy.

\section{Conclusion}
In this work, we introduced a Fusion-Guided Inception Network (FGIN) for hyperspectral single-image super-resolution. The proposed model integrates spectral-spatial fusion, Inception-like hierarchical feature extraction, and a multi-scale fusion mechanism to improve spatial feature learning while preserving spectral integrity. In addition, we designed an optimized upsampling module to retain fine details and ensure high-fidelity reconstruction. Experimental evaluations on two publicly available hyperspectral datasets demonstrate that FGIN achieves competitive performance while maintaining a relatively compact architecture. These results highlight the practicality and efficiency of FGIN for real-world hyperspectral super-resolution tasks.

\section*{Acknowledgment}
This project has been funded by the European Union’s NextGenerationEU instrument and the Research Council of Finland under grant \textnumero{} 348153, as part of the project \emph{Artificial Intelligence for Twinning the Diversity, Productivity and Spectral Signature of Forests} (ARTISDIG). We also gratefully acknowledge CSC for providing access to the LUMI supercomputer, operated by the EuroHPC Joint Undertaking.

\vspace{12pt}

\bibliographystyle{IEEEtran}
\bibliography{references.bib}

\end{document}